\documentclass{ieeeaccess}
\usepackage{cite}
\usepackage{amsmath,amssymb,amsfonts}
\usepackage{algorithmic}
\usepackage{graphicx}
\usepackage{textcomp}
\usepackage{color}
\usepackage{url}

\newcommand\hl[1]{\textcolor{black}{#1}}

\def\BibTeX{{\rm B\kern-.05em{\sc i\kern-.025em b}\kern-.08em
    T\kern-.1667em\lower.7ex\hbox{E}\kern-.125emX}}
\begin{document}
\history{Date of publication xxxx 00, 0000, date of current version xxxx 00, 0000.}
\doi{10.1109/ACCESS.2021.3072997}

\title{Breast Mass Detection with Faster R-CNN: On the Feasibility of Learning from Noisy Annotations}

\author{\uppercase{Sina Famouri}\authorrefmark{1},
\uppercase{Lia Morra\authorrefmark{2}, Leonardo Mangia\authorrefmark{3} and Fabrizio Lamberti}.\authorrefmark{4}}
% \IEEEmembership{Member, IEEE}}
\address{Address: Politecnico di Torino, Corso Duca degli Abruzzi, 24, 10129 Torino TO}

\address[1]{e-mail: sina.famouri@polito.it}
\address[2]{e-mail: lia.morra@polito.it}
\address[3]{e-mail: leonardo.mangia@studenti.polito.it}
\address[4]{e-mail: fabrizio.lamberti@polito.it}

% \tfootnote{This paragraph of the first footnote will contain support 
% information, including sponsor and financial support acknowledgment. For 
% example, ``This work was supported in part by the U.S. Department of 
% Commerce under Grant BS123456.''}

\markboth
{Famouri  \headeretal: Breast Mass Detectino with Faster R-CNN}
{Famouri \headeretal: Breast Mass Detectino with Faster R-CNN}

\corresp{Corresponding author: Sina Famouri (e-mail: sina.famouri@polito.it).}

\begin{abstract}
In this work we study the impact of noise on the training of object detection networks for the medical domain, and how it can be mitigated by improving the training procedure. Annotating large medical datasets for training data-hungry deep learning models is expensive and time consuming. Leveraging information that is already collected in clinical practice, in the form of text reports, bookmarks or lesion measurements would substantially reduce this cost. Obtaining precise lesion bounding boxes through automatic mining procedures, however, is difficult. We provide here a quantitative evaluation of the effect of bounding box coordinate noise on the performance of Faster R-CNN object detection networks for breast mass detection. Varying degrees of noise are simulated by randomly modifying the bounding boxes: in our experiments, bounding boxes could be enlarged up to six times the original size. The noise is injected in the CBIS-DDSM collection, a well curated public mammography dataset for which accurate lesion location is available. We show how, due to an imperfect matching between the ground truth and the network bounding box proposals, the noise is propagated during training and reduces the ability of the network to correctly classify lesions from background. When using the standard Intersection over Union criterion, the area under the FROC curve decreases by up to 9\%. A novel matching criterion is proposed to improve tolerance to noise. 
\end{abstract}


\titlepgskip=-15pt

\maketitle

\section{Introduction}
\label{sec:introduction}

\hl{In the last years, deep learning has led to major breakthroughs in many fields, including computer vision \cite{lecun2015deep}, medical imaging \cite{morra2019artificial,sahiner2019deep}, cyber-security \cite{xin2018machine,khosravy2020deep}, and many others. In medical imaging, the performance of deep learning systems often surpasses that of conventional machine learning systems and in some cases even rival that of experienced physicians \cite{sahiner2019deep}. However, one of the crucial ingredients of this success is the availability of large curated image collections on which deep models can be trained. As a matter of fact, data starvation} is often mentioned as one of the key obstacles to the application of deep learning in radiology \cite{kohli2017medical}. Nonetheless, this notion is only partially correct \cite{MDIMGsurvey}. Unlike other medical specialties, radiology departments are mostly digitized, and thousands if not millions of images are routinely stored in picture archiving and communication systems (PACS). Besides privacy concerns, a key obstacle to exploitation is the high cost of collecting annotations \cite{kohli2017medical,MDIMGsurvey}. For the evaluation of Computer Aided Detection/Diagnosis (CAD) systems, a lesion-level reference standard (or ground truth) is usually established by marking lesions on the image based on an independent gold standard (e.g., biopsy) or, when not available, a panel of expert radiologists, to account for high inter-rater variability \cite{petrick2013evaluation}. This strategy does not scale well to large datasets. 

Information about lesion location and characteristics are routinely collected on PACS and reading workstations by radiologists, in the form of textual reports, various kinds of bookmarks, and lesion measurements \cite{kohli2017medical,yan2018deeplesion,morra2015breast}. This information could be harnessed to automatically mine an approximate reference standard with limited additional costs. However, compared to human-annotated datasets, quality is usually compromised for quantity, and it is important to estimate the potential impact on \hl{the performance of deep neural networks trained on imperfect reference standard} \cite{irvin2019chexpert,yan2018deeplesion,tran2019distant}. In the future, a more widespread adoption of structured reporting is expected to further facilitate this practice \cite{kohli2017medical}. 

Some authors have used text mining to automatically extract labels from free-text reports \cite{irvin2019chexpert}.  Unfortunately, normally only patient-level labels can be generally obtained in this way; weakly supervised strategies such as Multiple Instance Learning (MIL) \cite{cheplygina2019not} can be used, but they usually perform poorly compared to fully supervised learning \cite{gao2018note}.

Recent works have shown the feasibility of establishing more precise and informative lesion-level annotations by mining hospital PACS \cite{yan2018deeplesion}. Reporting workstations commonly provide drawing tools such as bounding boxes (ellipses or squares), arrows, lines or diameters, that radiologists can use to bookmark and measure specific lesions \cite{yan2018deeplesion,morra2015breast}. A study conducted at the NIH Clinical Center found that the number of CT scans with such bookmarks skyrocketed after 2015; bookmarks often presented in the form of ellipses (8.4\%) or lesion diameters (46\%)\cite{yan2018deeplesion}. The recently released DeepLesion dataset, which contains over 32,000 lesions identified on CT images based on diameter measurements, shows the potential of this approach \cite{yan2018deeplesion}. Such annotations can be used to train object detection networks that, compared to image-level classifiers, can provide both lesion detection and localization \cite{ribli2018detecting,yan2018deeplesion}.

Mining strategies are attractive, but inevitably inject some levels of noise in the reference standard \cite{gao2018note,karimi2020deep}. For instance, there is no requirement that all the lesions mentioned in the report are explicitly annotated \cite{yan2018deeplesion}, and individual reports inevitably suffer from large inter-rater variability \cite{sacchetto2016mammographic,armato2011lung,kugler2018bad}. Research annotations are usually collected using two- or three-dimensional bounding boxes drawn as tight as possible to the lesion boundaries \cite{morra2015breast} or by segmenting the lesion \cite{armato2011lung}. Based on the authors' experience, bookmarks collected in clinical practice do not need to be as precise, and may serve additional purposes other than annotating the lesion (e.g., identifying the area selected for biopsy or further workup).

The present paper connects the problem of lesion detection from crowd-sourced annotations to the effect of noisy annotations on the generalization error of deep neural networks, and in particular object detection networks. In this context, previous studies have focused on changes in the ground truth labels (e.g., missing or mislabelled objects) \cite{gao2018note}. Here, we identify a different and independent source of noise, which results from loosely annotated lesions (i.e., bounding boxes are approximate and include the lesion as well as part of the background). We show how this source of noise can potentially affect the detection performance, and investigate its implications by performing controlled experiments in which an increasing amount of noise is injected.

Many different architectures have been proposed for object detection \cite{frcnn,lin2017focal,redmon2016you} and successfully applied to medical imaging \cite{ribli2018detecting,wu2019deep,agarwal2019automatic,cha2019reducing,jung2018detection,al2018simultaneous}. However, all architectures share some common operating principles: they include one or more classification modules that classify Regions of Interest (ROIs), each identified by a bounding box, as either background or one of the possible object (lesion) classes. These modules are trained by selecting examples of bounding boxes labeled as foreground (positive examples) or background (negative examples). \hl{Discrimination of positive and negative examples is achieved} by comparing object proposals with the ground truth bounding boxes based on a \textit{matching criterion}: usually, a threshold is set on the Intersection over Union (IoU) to determine if two boxes are a match. If the ground truth is imprecise (e.g., the bounding boxes are larger than the actual object), the matching may be incorrect, hence the classification modules will be trained on noisy labels and detection performance may suffer. It is precisely this phenomenon that we seek here to quantify and characterize. 

Our experiments are conducted on the CBIS-DDSM (Curated Breast Imaging Subset of DDSM) dataset, a public screen-film mammography dataset for which high quality, clean labels are provided. In order to conduct a controlled experiment, noise is artificially injected by varying the size of the bounding boxes. Our reference architecture is the Faster R-CNN, which was shown to perform quite well for breast mass detection \cite{ribli2018detecting,cha2019reducing}. It must be stressed, however, that matching object proposals with the ground truth is a common step for all object detectors and, while different detectors generate object proposals in different ways, matching is usually performed based on the IoU.

We establish the performance of the network in the presence of different noise levels and different matching criteria. When the bounding boxes increase in size, the number of ROIs labeled as positive increases, which is likely due to background being incorrectly labeled as foreground. Despite this fact, our experimental results show that object detection is quite  robust in the presence of low to moderate amount of noise. In the presence of moderate to large noise, a simple yet effective countermeasure consists in the use of alternative matching criteria, favoring examples that are closer to the center of the ground truth bounding box and, thus, more likely to contain the actual lesion. Training lesion detectors by using imprecise bounding boxes is in principle feasible and, by carefully setting up the training procedure, even robust to high levels of noise. 

The rest of the paper is organized as follows. Related works and background on Faster R-CNN are presented in Section \ref{related}. Explanation of the noise model, network training, and the matching criteria used in this work are given in Section \ref{methods}. Experimental setup and results are reported in Section \ref{sec:results}, and discussed in Section \ref{discussion}.

\section{Related Work}
\label{related}

\subsection{Object detection and Faster R-CNN} \label{fR-CNNexp}

Object detectors operate by classifying potential object regions that can be generated using fixed grids (one-stage detectors) or by employing a pre-selection mechanism (two-stage detectors). The Faster R-CNN \cite{frcnn} is a well-known two-stage architecture for object detection, whereas single-stage architectures include YOLO \cite{redmon2016you} and RetinaNet \cite{lin2017focal}. 

Faster R-CNN is composed of two modules: the Region Proposal Network (RPN) and the detector. Both networks share the same convolutional backbone for feature extraction, which is usually pre-trained for classification and fine-tuned for object detection. Both the RPN and the detector have two \textit{heads}, one for predicting the bounding box (regression head) and one for the classification.  

The RPN takes an image (of any size) as input and outputs a set of object proposals or ROIs; it performs a binary classification to separate objects (of any class) from the background and narrow down the search. In order to do so, a sliding window is passed over the feature map and, at each location, $k$ proposals with different scales and aspect ratios, known as \textit{anchor boxes}, are generated (a common choice is $k=9$). The RPN takes as input the coordinates of each anchor box, along with the feature map, and predicts the coordinates of the object bounding boxes, along with a binary score. 

The RPN commonly predicts multiple overlapping bounding boxes for the same object. To further reduce the number of proposals, Non-Maximum Suppression (NMS) is used to reduce the number of proposals passed to the classifier: in short, for each group of overlapping bounding boxes, only the box with the highest classification score is retained, whereas the others are discarded. The detector then calculates the final classification score and the final bounding box, taking as input the feature map and the RPN proposals. 

Training of the two heads is performed jointly in an alternating fashion. At each forward pass (where a forward pass corresponds to one image), the RPN is trained and updated; then, the output of the RPN is kept fixed and the detector head is updated. The loss is defined for both modules as a combination of a regression and classification loss \cite{frcnn}:

\begin{multline}
    L(\{p_i\}, \{b_i\}) = \frac{1}{n_c}\sum_{i}L_{cls}(p_i, p'_{i}) + \\ + \lambda\frac{1}{n_r}\sum_{i}{p_i * L_{reg}(b_i, b'_{i})}
\end{multline}

\noindent where  $L_{reg}$ is the smooth L1 loss for regression,  $L_{cls}$ is the categorical cross entropy, $b_i$ is the ground truth bounding box, $b'_i$ are the output coordinates, $p'_{i}$ is the predicted probability that $b_i$ contains a lesion, and $p_i$ is the ground truth label. This loss is used for both the RPN and the detector.

During training, each anchor box is labeled as a positive, negative or neutral example based on its overlap with the ground truth annotations. In the original Faster R-CNN architecture the overlap is calculated based on the IoU. Neutral examples, which are usually borderline (e.g., partially overlapping), are not used during training. The same process is repeated to train the detector. As it will be shown in Section \ref{sec:matching}, the results critically depend on the definition of the matching criterion. 

\subsection{Labeling Noise in Deep Neural Networks}
\label{sec:noise}

\color{black}
Real-world datasets are often affected by several forms of noise, which may prevent a machine learning model to correctly identify patterns in data. This is especially true in medical imaging, since accurate labels are expensive to obtain, pathological signs are often ambiguous, and inter-observer variability is high  \cite{karimi2020deep,sacchetto2016mammographic}. The notion of label noise is not conclusive in the literature, where the term has been used to refer to different forms of label imperfections or corruption\cite{karimi2020deep}.  

To the best of our knowledge, the effect of noise in training object detectors is relatively unexplored in literature, especially in the medical domain; compared to the general problem of object detection, lesions have much more ill-defined margins and are much rarer, leading to an extremely high-class imbalance. Therefore, the closest works in literature are those related to classification labelling noise, which has been extensively studied \cite{gao2018note,frenay2014classification,reed2014training,Benoit2014IntroLabeNoise,LabelsCorruptedHendrycs}. Given a set of training samples $ \{x_i, y_i\}$, where $y_i$ is a discrete variable that corresponds to the true class of the sample, labelling noise can be formally defined as a stochastic process which pollutes the labels that are passed to the learning algorithm \cite{Benoit2014IntroLabeNoise}. As a result, the observed label may no longer represent the true class of an instance.

Many noise models have been proposed for classification problems. The most common transformations studied are label flips and outliers \cite{noisycnn}. Label flips are referred to samples that have been given a wrong class label, whereas outliers are samples which do not belong to any of the classes in the training set. Another important distinction is whether the noise affects all classes uniformly, or whether the noise is statistically dependent on the class or the features \cite{Benoit2014IntroLabeNoise}. For instance, in the medical domain, negative or borderline results may be affected by higher noise since biopsy is not performed. We do not consider here the case in which label noise is constructed adversarially. 

Various works have sought to establish the effect of labeling noise on classification performance in various regimes, either from a theoretical or experimental standpoint, and often reaching diverging conclusions. On the one hand, theoretical results \cite{natarajan2013learning} imply that a high capacity model should be robust to several types of random noise, provided that a sufficient number of training samples is available. To some extent, deep learning methods have indeed shown resilience to labeling noise given a required number of clean labels. For example, experiments on MNIST in \cite{RobustmasslabelnoiseRolnick} showed that with a 10:1 ratio of noisy to clean labels, at least 2000 clean labels are needed to reach an accuracy of 90\%, whereas for a ratio of 50:1, the number of clean labels required increase to 10,000 for reaching the same performance. However, in practice the number of training samples is usually limited, especially for medical applications, and the label noise may follow complex, class-dependent patterns. Due to their memorization effect, sooner or later deep neural networks start to memorize noisy labeled samples, especially if the percentage of noisy labels is relatively large \cite{zhang2016understanding, co-teaching}. 

It is therefore of practical interest to investigate experimentally the effect of noise in different tasks/datasets, in order to establish the feasibility of training deep neural networks \cite{kugler2018bad,xue2019robust}.  To test the robustness of algorithms against noise under controlled conditions, a common strategy is to inject synthetically generated noise into a clean gold standard dataset, progressively increasing the percentage of noisy labels \cite{noisycnn,dgani2018training,xue2019robust}. While the most common strategy is to randomly flip a fraction of the labels, either based uniform or class-dependent random distribution, the generated noise is not always realistic, and other problem-specific strategies may be useful. For instance, Xue et al. simulated labelling noise in skin lesion classification by training a DNN on a fraction of the dataset, sorting the remaining images according to the test loss, and finally flipping the top $x$\% of the high loss samples for both classes \cite{xue2019robust}. This method accounts for the fact that mislabelling by dermatologists does not depend on the class, but rather on the training sample, and is most likely to occur with difficult or borderline cases. They demonstrated a 20\% decrease in accuracy when the noise ratio increased up to 40\% \cite{xue2019robust}.  

In our work, we focus on object detection networks, and specifically on box coordinate noise (which we will refer to as \textit{noise} in the remainder of this paper), assuming that the class label is \textit{per se} correct. To the best of our knowledge, this type of noise has not been investigated yet. As we will show in our experiments, coordinate noise can translate into class label noise when training the RPN and detector classifier heads. In fact, as introduced in Section \ref{fR-CNNexp}, the training procedure labels anchor boxes as positive, negative, or neutral examples using a matching criterion. We will show how box noise can introduce labelling noise as a mix of class label flips (e.g., a positive anchor box becomes negative or viceversa) or outliers (e.g., a neutral bounding box becomes positive or negative).

Besides studying the practical effect of labelling noise, many authors have explored methods to make training  \textit{robust}   in the presence of noise \cite{frenay2014classification,xue2019robust}, either by suggesting noise-tolerant losses \cite{ghosh2017robust,reed2014training,sukhbaatar2014dvggfdstraining,patrini2017making}, or by detecting and removing noise from the dataset through dedicated networks or models \cite{co-teaching,noisycnn,dgani2018training}. The interested reader is referred to recent surveys for a more in-depth discussion on this subject\cite{karimi2020deep}. 

In the field of object detection, previous work by Gao and colleagues showed how mining procedures, which are used to complete a partially annotated dataset, may introduce several types of labelling noise, such as false negatives, false positives and box coordinate noise \cite{gao2018note}. To counter the effect of each type of noise, they added an ensemble classification head and a distillation head \cite{distilling} in order to avoid overfitting to the labeling noise, and modified the loss to compensate the effect of false negatives. However, they trained the box regression heads only on seed annotations to circumvent the effect of box coordinate noise. Our work is different since we concentrate specifically on this type of noise, and show that it can affect the training not only of the regression, but also of the classification heads. We also show that the matching criterion plays an important role in making the object detector more robust against noise, and propose ways to mitigate performance degrading.  
\color{black}

\subsection{Breast Mass Detection}\label{bmdetection}

CNN-based models are currently the state of the art in breast image analysis, in some cases reaching performance closes to those of human experts \cite{kooi,levy2016breast,ribli2018detecting,wu2019deep,schaffter2020evaluation}. 

Deep learning-based approaches to breast lesion detection and localization fall into two main approaches based on whether case-level or lesion-level annotations are available. At the case-level, recent works have shown that multi-stream CNNs can estimate the presence or absence of malignant findings from multi-view mammography images \cite{wu2019deep,mckinney2020international}. Such experiments relied on large scale datasets, ranging from 100,000 to 200,000 cases and more than 1,000,000 images. 

Other works have exploited CNN-based object detectors trained on lesion-level annotations, e.g., in the form of bounding boxes drawn around the lesion \cite{ribli2018detecting,cha2019reducing,mckinney2020international,schaffter2020evaluation,agarwal2019automatic,jung2018detection,al2018simultaneous}. From the radiologist perspective, this approach has the advantage of providing precise localization of cancerous lesions, which enables the method to be directly used as a CAD tool. By leveraging information about lesion localization during training, object dectors also have the potential to exploit training data more effectively \cite{schaffter2020evaluation}. Indeed, within the DREAM challenge \cite{DREAM}, a Faster R-CNN model by Ribli et al. \cite{ribli2018detecting} reached the second position in the competitive phase, and scored first in the subsequent collaborative phase when more training data became available \cite{schaffter2020evaluation}. It should be noticed, however, that in the DREAM challenge itself, only case-level annotations were available, and the model had to be trained on separate private datasets, in part collected by the research group \cite{schaffter2020evaluation}. This fact highlights both the benefits and challenges of acquiring lesion-level annotations at large scale, since precise annotations have to be manually added by an experienced radiologist for research purposes. The present work aims at establishing whether these requirements can be relaxed, paving the way for future crowd-sourcing of annotations. 

Several architectures were proposed for object detection. Compared to general-purpose object detection benchmarks, medical applications place a higher emphasis on accuracy than execution speed, as real-time performance is usually not needed. This is evident in mammography applications which employed architectures renowned for their accuracy, such as Faster R-CNN \cite{ribli2018detecting,cha2019reducing,schaffter2020evaluation,agarwal2019automatic} and RetinaNet \cite{mckinney2020international,jung2018detection}, with few exceptions based on YOLO \cite{al2018simultaneous}. On the public InBreast dataset \cite{moreira2012inbreast}, solutions based on Faster R-CNN outperformed those based on RetinaNet, achieving 92\% sensitivity at 0.3 False Positive (FP)/image \cite{agarwal2019automatic}. However, many differences in the experimental settings hinder direct comparison between papers: for instance, since the InBreast dataset is small, research groups relied on different public and private digital mammography datasets in order to build the training sets.

In our experiments, for the sake of reproducibility we exploited the publicly available CBIS-DDSM dataset, choosing the Faster R-CNN architecture based on the above considerations. Hyper-parameters of the Faster R-CNN architecture were tailored to the specific challenges of mammography, also taking into account previous works \cite{ribli2018detecting,cha2019evaluation}. We found in particular that Faster R-CNN overfits the CBIS-DDSM dataset, which is consistent with previous experiments by \cite{cha2019reducing}. We also found that the choice of proposals during training has an important impact on generalization; thus, we propose to use hard sample mining on region proposals to increase performance (Section \ref{model}).

\begin{figure*}[!htb]
    \centering
    \includegraphics[width = 1\textwidth]{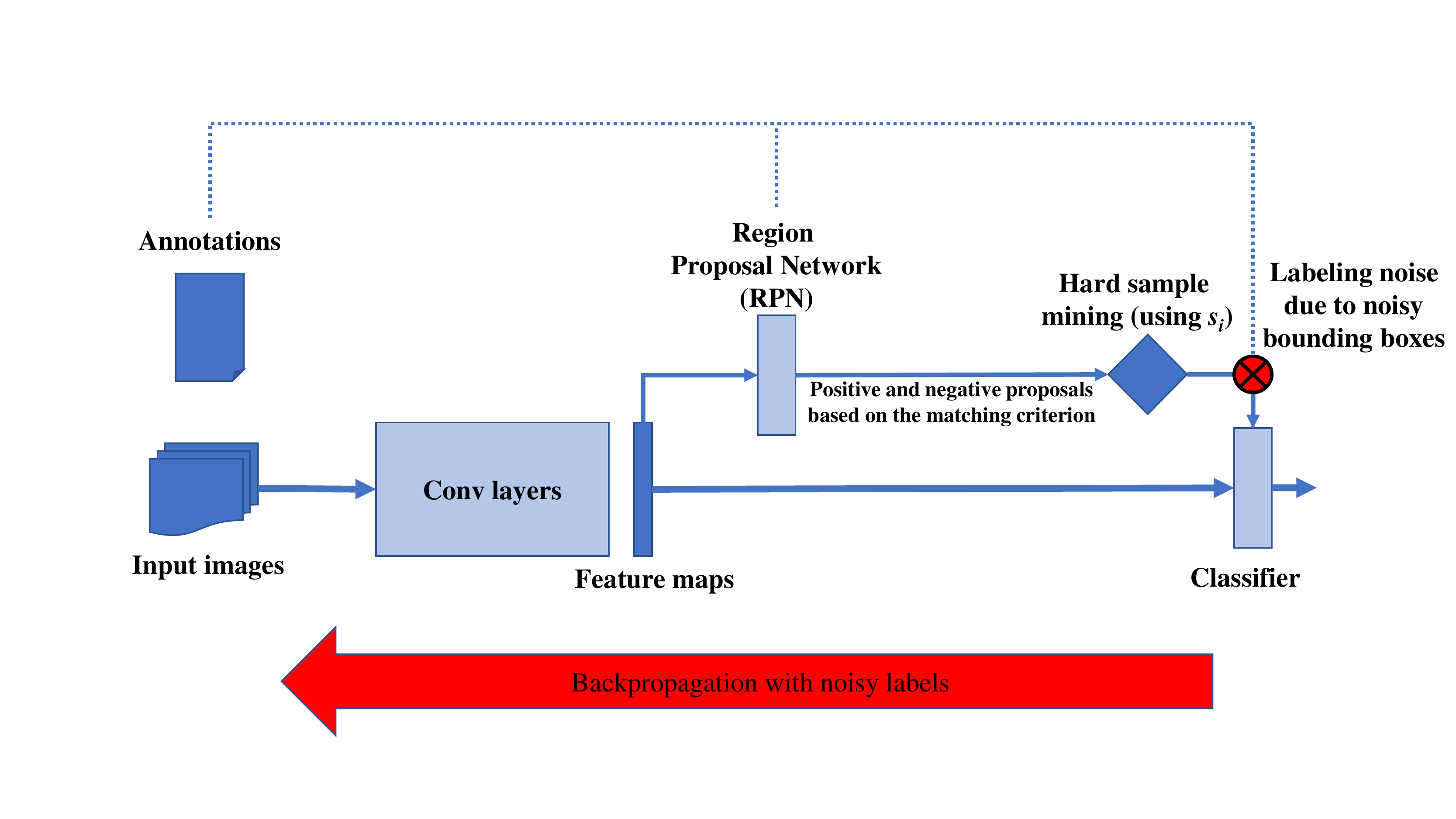}
    \caption{\hl{The training framework of Faster R-CNN is summarized in this figure, where the noise is injected on the bounding boxes in the annotations and its effect on the training is pinpointed by the red color. The red circle shows where the labeling noise is generated. Faster R-CNN is a two-stage object detector: the RPN filters candidate regions from the background, and the classifier assigns them the most likely class. Since lesions are rare compared to the background, a hard sampling procedure has been added to the framework to avoid overfitting and make sure that more informative proposals are passed to the classifier. If the reference bounding boxes are not tight to the lesions, the proposals that are passed to the classifier may contain noisy labels. In fact, proposals are automatically labelled against the reference standard on the basis of a matching criterion, such as the IoU, which may lead to incorrect results if the bounding boxes are not tight to the lesions.}}
    \label{training_summary}
\end{figure*}

\subsection{Metrics and losses for Bounding Box regression}
\label{sec:metrics}

A common problem of object detection and CAD systems is how to evaluate whether, and to what extent, an object proposal corresponds to a given ground truth bounding box. The IoU is the most common evaluation metric in object detection \cite{frcnn,redmon2016you,lin2017focal,rezatofighi2019generalized}. Its value, bounded between 0 (no overlap) and 1 (complete overlap), measures how tightly the detector output fits the ground truth bounding boxes. As introduced in Section \ref{fR-CNNexp}, the quality of the bounding box depends mostly on the regression head. 

Recently, two variants of the IoU have been proposed with the purpose of building more robust regression losses \cite{rezatofighi2019generalized,zheng2019distance}. Since the IoU is \emph{per se} only defined when two bounding boxes overlap and does not provide any gradient for the non-overlapping cases, it cannot be used directly as loss for the regression head. Usually, a surrogate loss is used, such as the smooth  $\mathit{l_1}$-norm or, more generally, a  $\mathit{l_p}$-norm \cite{frcnn}. The Generalized Intersection over Union (GIoU) \cite{rezatofighi2019generalized} yields non-zero values for non-overlapping bounding boxes, thus overcoming one of the main limitations of the IoU. The Distance-IoU \cite{zheng2019distance} combines the IoU with the Euclidean distance between the central points of the proposal and ground truth bounding boxes. Both metrics were used to define regression losses that could be incorporated in existing architectures, with performance gains especially for fast architectures such as YOLO \cite{zheng2019distance,rezatofighi2019generalized}. 
In the present research, the focus is on improving the performance of the classifier, rather than the regression head. Furthermore, our experimental settings assume that the bounding boxes in the ground truth have a systematic bias, which would be certainly captured by the regression parameters.

In this respect, the IoU metric still plays an important role, as it is used to label ROIs during training. In all detectors, ROI proposals that overlap a ground truth bounding box with an IoU greater than a certain threshold, such as 0.5 \cite{lin2017focal} or 0.7 \cite{frcnn}, are marked as foreground, or true positives (TPs), whereas the rest are labeled as background, or false positives (FPs). Both \cite{rezatofighi2019generalized} and \cite{zheng2019distance} assume that the ground truth is noise-free, and do not investigate the properties of IoU, GIoU and Distance-IoU in the presence of noise. 

When a bounding box proposal does not overlap with the ground truth, it can be safely labeled as background. Hence, the GIoU metric \cite{rezatofighi2019generalized} is not particularly relevant in this context, since the main difference with the IoU is in the case of non-overlapping bounding boxes. On the other hand, in the presence of noise the number of proposals that will overlap with the ground truth is expected to increase significantly, and matching criteria that incorporate both distance and overlap measures (e.g., \cite{zheng2019distance}) may be able to better differentiate between competing overlapping proposals. In \cite{zheng2019distance}, however, this aspect was not investigated. 

The present work also draws inspiration from the CAD evaluation field: here, compared to the field of object detection, a much wider range of mark-labeling or matching criterion have been proposed to decide which CAD marks correspond to the targeted abnormalities \cite{petrick2013evaluation}. Mark-labeling rules are usually defined based on a measure of overlap, distance, or a combination them, also depending on the type of graphical marks used to represent the CAD output and/or to record the ground truth. Another common criterion is verifying whether the centroid of the CAD mark falls inside the ground truth bounding box \cite{ribli2018detecting}. Mark-labeling rules commonly used in object detection benchmarks may be too strict for CAD applications. For the clinical purposes of a CAD system,  there is less interest on the tightness of the predicted bounding boxes, since lesions are sparse, often with ill-defined margins, and occlusions not existent. Previous research established that different criteria can result in dramatically different TP and FP estimates \cite{petrick2013evaluation}. In this paper, different matching criteria are evaluated and compared in the presence of noise.

\section{Materials and methods}
\label{methods}

\hl{The training framework for Faster R-CNN contains several modules that are summarized in Figure \ref{training_summary}. Starting from the left, the input images alongside the annotations are passed to the network. The annotations are used to localize and identify the lesion and its type. Accordingly, during training, the annotations are considered as ground truth to calculate the loss and then via backpropagation the weights will be updated. However, if the bounding boxes are noisy (i.e., not tight to the lesions), we argue that they would lead to labeling noise when the region proposals are passed to the classifier, as discussed in Section \ref{sec:matching}. Backpropagation with noisy labels will result in lower performance in lesion detection. Therefore, we set up a framework to analyse this issue and provide a solution to have more robustness against such noise. The rest of this section is dedicated to explaining each module in detail.} 

\subsection{Dataset}

The CBIS-DDSM collection \cite{lee2017curated,LeeTCIA,clark2013cancer} is an updated version of part of the DDSM dataset \cite{ddsm}, selected and curated by a trained mammographer, and available for download from the Cancer Imaging Archive (TCIA). DDSM contains 2620 scanned screen film mammography studies, including normal, benign, and malignant findings  with verified pathology information. Each study is composed of up to four images acquired in the cranio-caudal (CC) and medio-lateral oblique (MLO) orientations; however, only images with findings are included in the CBIS-DDSM dataset, for a total of 3,089 images. The CBIS-DDSM database has been preprocessed using standard techniques for screen-film mammography and converted to DICOM, as detailed in \cite{LeeTCIA}. Additionally, we cropped the breast region before processing using a previously developed automatic algorithm \cite{morra2015breast}, with the sole purpose of reducing computational time. In this study, we focus only on the detection of \hl{mass findings}, for which accurate segmentation \hl{by experienced radiologists} is available in the CBIS-DDSM collection. \hl{The ground truth bounding boxes are defined as the tightest bounding box that completely encloses the segmentation: we can therefore assume that the initial bounding boxes are very precise and essentially noise free.} For microcalcification clusters, only a coarse segmentation is provided. Hence, we postulate that the type of noise we wish to study can be more accurately modelled for masses, and excluded such cases from the dataset. We used the standard training/test split (80\%/20\%) defined by the CBIS-DDSM authors.\hl{ The final training and test set included 550 patients (with 613 masses) and 200 patients (with 222 masses), respectively. For each patient, up to four images are available and, since each finding is in most cases visible in both CC and MLO views, the total number of lesion views is 1,316 and 374 for the training and test set, respectively.}

\subsection{Noise Modeling}

Let us assume that a clean ground truth bounding box $b_i = (x_{1i}, y_{1i}, x_{2i}, y_{2i})$ is available for lesion $i\in\{1,2,...,m\}$, and that it perfectly fits the lesion boundaries. In order to inject noise, each $b_i$ was modified by a random factor (noise), defined as:

\begin{equation}
    \begin{split}
        w'_i = (1+n_{wi})w_i, \\
        h'_i = (1+n_{hi})h_i.
    \end{split}
\end{equation}

\noindent where  $(w_i, h_i)$ are the width and height of $b_i$, $(w'_i, h'_i)$ are the width and height of $b'_i$, and $(n_{wi}, n_{hi})$ are sampled from a normal distribution $n_{wi}, n_{hi} \sim \mathcal{N}(\mu,\,1)$ with mean $\mu$. Bounding boxes are defined in pixels in the ground truth, whereas $\mu$ is defined as a dimensionless multiplicative noise.

In practice, since the purpose of the bounding box would be to roughly bookmark the location of the region, we assume that the bounding box $b'_i$ is always equal or larger than the clean bounding box, and that extremely large bounding boxes are likewise unlikely. Hence, we clipped $n_{wi}$ and $n_{hi}$ in the range [0,6), i.e., we assume the final bounding box is at most six times larger than the original. 

Despite this limit, since the typical size of mammography masses ranges between 1 and 3 cm, the resulting bounding box may still exceed the breast region, which is unrealistic: based on the size of the largest lesions in CBIS-DDSM, we truncated $b'_i$ to be at most 80\% of the total image width. The center of $b'_i$ is as same as $b_i$, except in those cases where the bounding box has been cropped to fit within these limits.

\begin{figure}[!htb]
    \centering
    \includegraphics[width = 0.45\textwidth]{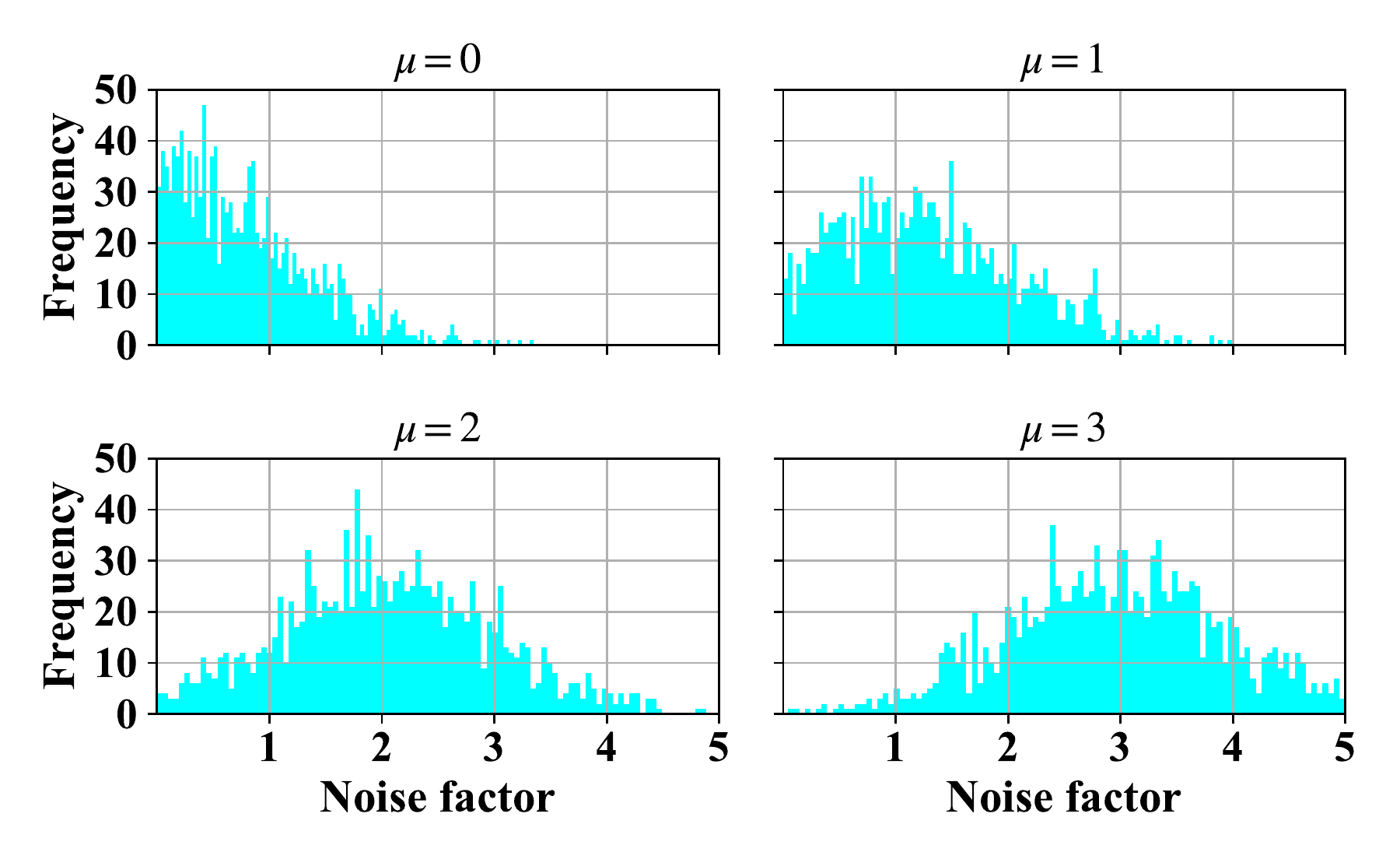}
    \caption{Histograms of the noise factor $n_{wi}$ from $\mu=0$ to  $\mu=3$. All distributions are clipped to the range [0, 5].}
    \label{noise_model}
\end{figure}

\begin{figure}[!htb]
    \centering
    \includegraphics[width = 0.45\textwidth]{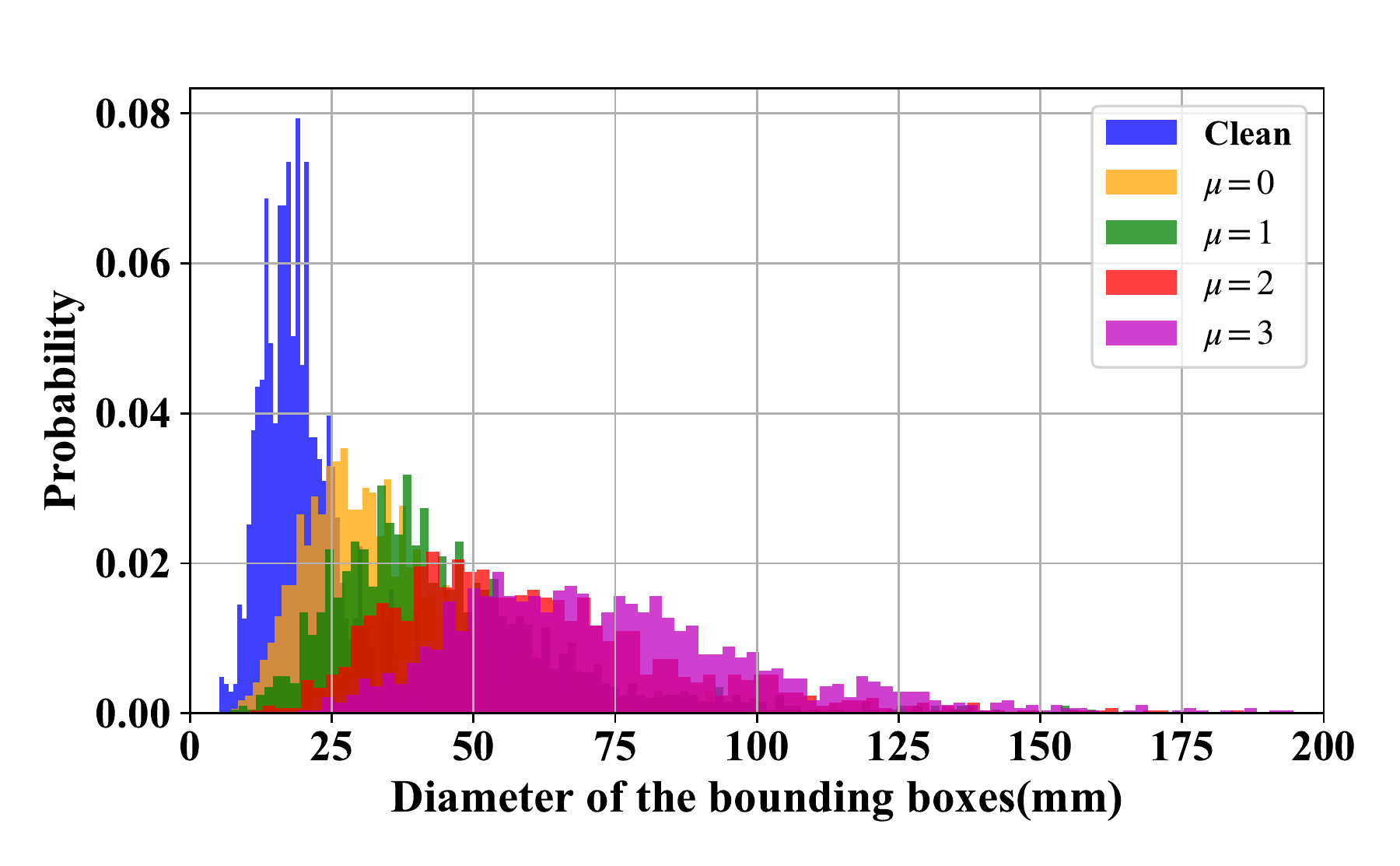}
    \caption{Diameter distribution (histogram) of the noisy bounding boxes for the clean dataset and for each level of noise.} 
    \label{diam}
\end{figure}

Four different levels of noise were generated with  $\mu=\{0,1,2,3\}$. Histograms of $n_{wi}$ and $n_{hi}$ drawn from the model are depicted in Figure \ref{noise_model}, whereas the distributions of the bounding boxes diameters at different levels of noise are compared in Figure \ref{diam}.

\subsection{Matching Criterion}
\label{sec:matching}

As anticipated in Section \ref{related}, a matching criterion is needed while training and testing to establish whether a bounding box proposal is a TP or FP detection. \hl{In other words, the matching criterion is a function that measures how well the ground truth bounding boxes, defined manually by experts, match the proposals given by the network.} We included in the experiments three matching criteria, drawing from established papers in CAD evaluation and object detection: the IoU, which is the \emph{de facto} standard in object detection, a simple Centroid-based criterion, and a combination of distance and overlap, which is denoted as  Exp\_IoU in the following. 

Anchor boxes with a IoU higher than a threshold $T_u$ are labeled as positive, and those lower than a second threshold $T_l$ as negative. The rest are ignored at training time (neutral examples). In the rare cases in which no bounding box can be labeled as positive based on the IoU threshold, the anchor with the highest IoU overlap is selected \cite{frcnn}. As previously done in \cite{ribli2018detecting}, we decrease the $T_u$ from the original 0.7 value to 0.5 in order to allow more positive examples in each batch. We did not experiment with higher thresholds, as they were previously found to lead to unstable training \cite{ribli2018detecting}. $T_l$ is instead equal to 0.3 \cite{frcnn}.

The Centroid-based criterion or ``centroid inside the bounding box'' simply checks whether the center of the proposed bounding box falls inside the ground truth bounding box. This is a common criteria for evaluating CAD systems, but has never been used for training \cite{ribli2018detecting}.

\begin{figure}[!htb]
    \centering
    \includegraphics[width = 0.45\textwidth]{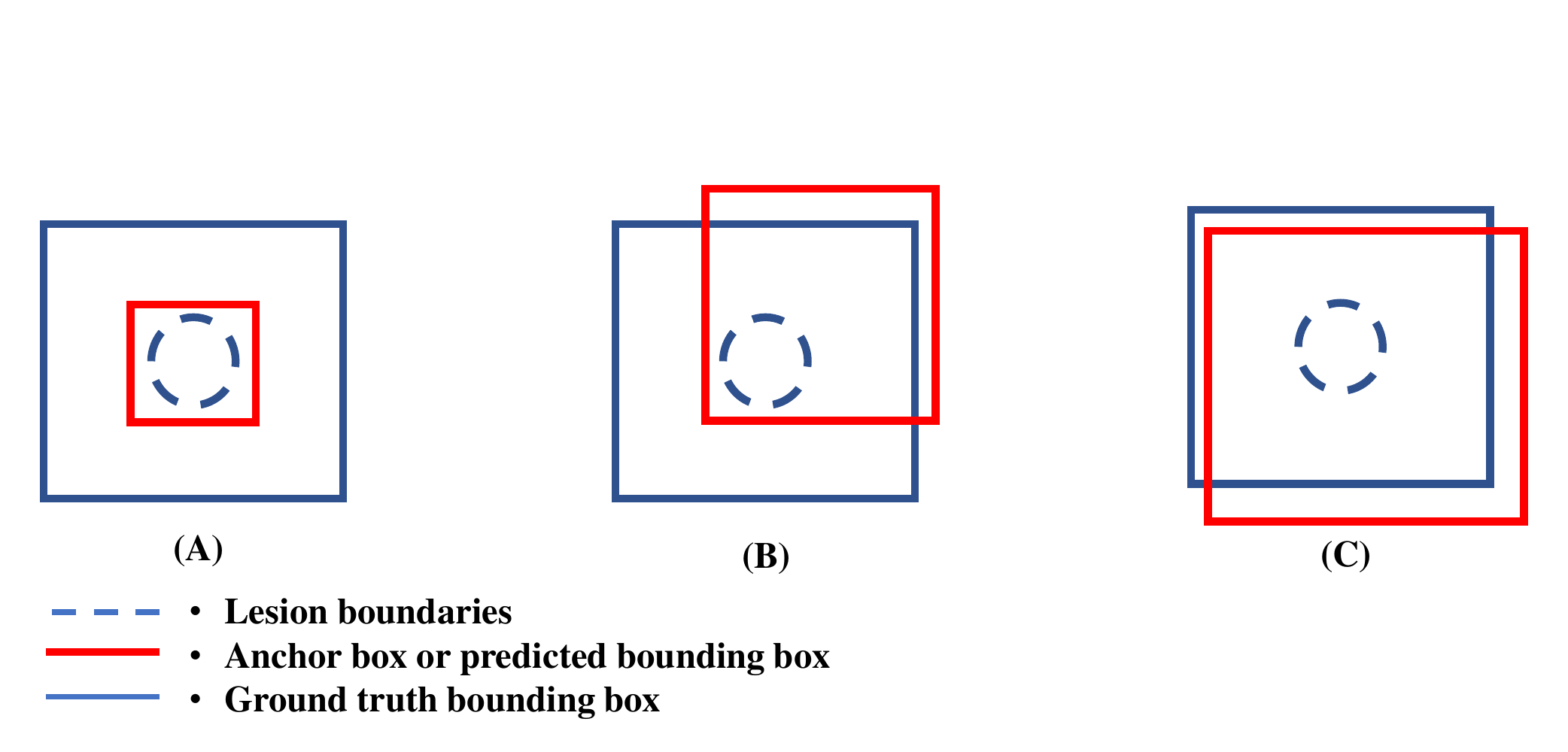}
    \caption{The red boxes show three different bounding box proposals for a lesion. If IoU is considered as the matching criterion, the order of the scores would be $C>B>A$. However, based on the distance between centroids, the order would be $A>C>B$. }
    \label{match_example}
\end{figure}

Overlap-based and distance-based metrics may produce different labels for a given bounding box. A simple example is given in Figure \ref{match_example}, which compares the ranking of proposed bounding boxes with respect to different criteria. Each criterion has different strengths: the IoU favors bounding boxes that match in both size and position, but may fail in the presence of noise (e.g., it may fail to match bounding boxes that are correct, but smaller, as in Figure \ref{match_example}A). Relying solely on the centroid is too weak, an intuition which is confirmed by experimental results in Section \ref{sec:results}.

Hence, we propose a new criterion, denoted Exp\_IoU, which explicitly considers both the size and relative position of the bounding boxes, as follows:

\begin{equation}
    S_{exp\_iou}(b'_i, b_j) = \frac{IoU(b'_i, b_j)+e^{-\beta D(b'_i, b_j)}}{2}
    \label{eq:exp_io}
\end{equation}

\noindent  where $b'_i$ is the proposed bounding/anchor box, $b_j$ corresponds to the ground truth, $D(.)$ represents the Euclidean distance between the centers of each bounding box, and $\beta$ balances the two contributions. The value of $\beta$ was set to $0.1$ based on experimental results. The same thresholds $T_u$ and  $T_l$ are used for both IoU and Exp\_IoU.

\subsection{Deep network architecture}
\label{model}

As said, the architecture is Faster R-CNN, which was introduced in Section \ref{fR-CNNexp}. We here explain how the network was parameterized, and present some enhancements that were introduced to cope with the unique characteristics of the medical domain. 

For the backbone network we chose the ResNet50 architecture, since preliminary investigations reported better results than VGG16 \cite{ribli2018detecting}. More specifically, layers up to conv4\_x are included in the backbone, and the top layers (conv5\_x) are included in the RPN and classifier heads. 

The anchor box scales are $\{128, 256, 512\}$, whereas the aspect ratios are $\{(1,1), (0.7, 1.4), (1.4, 0.7)\}$, with a stride of 16. Therefore, at each location nine anchor boxes will be generated. The NMS overlap threshold is set to 0.7 for training and 0.1 for testing; this threshold is much lower than in the original paper, but is justified by the fact that lesion margins in mammography are much less defined than object boundaries in natural scenes \cite{ribli2018detecting}. NMS limits the number of generated bounding boxes at 300 at both training and testing time. The value of $\lambda$ is set to $8.3$ for the RPN loss, and $12.5$ for the detector.

The RPN generates at most 300 examples per image. This results in a severely imbalanced classification problem, since the background is even more predominant than in general object detection. In fact, even considering only images which contain at least one lesion, the number of positive samples generated by the RPN is usually between 2 and 10. A possible strategy is to randomly sample a balanced subset: in our experiments, we chose four random ROIs (two positives and two negatives). This however may result in a poor selection of samples, and thus impair the network training.

A simple heuristic was added to mine informative examples after NMS. We assume that hard samples are those misclassified by the RPN, since one of the main difficulties in mammography is separating lesions from glandular patterns that mimic their presence. The following score is used to rank all the proposed ROIs:

\begin{equation}
    s_i = (p^{'}_{i} - p_i)^2
\end{equation}

where $p^{'}_{i}$ is the predicted probability that $b_i$ contains a lesion, and $p_i$ is the ground truth label. As $s_i$ increases, the margin between the probability and the true label grows, meaning that it is a hard sample. In the presence of labeling noise, the margin $s_i$ is also expected to be higher for noisy samples than clean ones \cite{gao2018note}.
 
The positive and negative samples are sorted separately based on their score $s_i$, and the corresponding mean scores $S_P$ and $S_N$ are calculated. The samples can then be split into four categories:

\begin{itemize}
    \item Easy positive: positive samples with $s_i < S_P$;
    \item Hard positive: positive samples with $s_i \geq S_P$;
    \item Easy negative: negative samples with $s_i < S_N$;
    \item Hard negative: negative samples with $s_i \geq S_N$.
\end{itemize}

From each category, 25 positive samples and 25 negative samples are selected in order to maintain a balance between difficult and easy examples, as well as noisy and clean ones. Finally, four ROIs are randomly sampled from this subset.

\subsection{Experimental Setup}

We performed a preliminary experiment in which we compared the baseline model (IoU matching criterion) with and without the hard negative sampling strategy. Then, performance across all levels of noise and matching criteria were compared, for a total of 15 possible configurations. All the experiments used the same hyper-parameters as explained in Section \ref{model}. The thresholds for IoU and the Exp\_IoU are defined in Section \ref{sec:matching}. Weights pre-trained on ImageNet were used to initialize the backbone.

The Adam \cite{adam} optimizer was used with a learning rate of $10^{-5}$; each network was trained for 80 epochs, each comprising 500 iterations. Experimentally, we observed that Faster R-CNN starts to overfit after this number of epochs. Images were downsampled so that the largest dimension was equal to 600 pixels to reduce the computational effort, even though if better results could be obtained with higher resolution images \cite{ribli2018detecting}. 

The network was implemented in Keras 2.2 with Tensorflow 1.13.1. In order to reduce the variability between different experiments, a constant seed was set for all libraries (Numpy and Tensorflow). The order of the images was randomized, but fixed for all experiments. All experiments were conducted on an Nvidia Titan Xp GPU.

\subsection{Evaluation}

All experiments were evaluated on the clean test set, without inserting noise. The Free-Response ROC (FROC) was used to plot sensitivity vs. the average number of false positives (FPs)/image \cite{afroc,petrick2013evaluation}. \hl{The FROC paradigm is a location-specific variant of ROC analysis where the number of detections per image is not constrained, and each detection can be assigned a separate score by the CAD algorithm. The FROC curve is thus more suited to evaluate object detection networks, and is a widely accepted methodology in the medical image analysis literature \cite{petrick2013evaluation}. }
\hl{The centroid inside the bounding box was used as the matching criterion for evaluation purposes. This is a commonly used methodology for evaluating CAD systems, and has been employed in similar works \cite{ribli2018detecting}. In fact, lesions are more sparse and have much less well-defined boundaries than objects on traditional images, and therefore tightly matching the reference bounding box is less relevant as long as the lesion is clearly shown to the radiologist. Furthermore, this choice allows us to compare performance at different noise levels on fair grounds. In fact, when comparing the predictions of networks trained on noisy bounding boxes against the clean ground truth, the IoU would significantly drop as the predicted bounding box is larger than the true lesion. Instead, we focus on whether each network can correctly locate the true lesion. }

The FROC curves were computed on 1000 bootstrap samples, each containing 200 cases (which is the size of the validation set). All FROC curves were cut at 2 FPs/image, as high false positive rates are clinically less useful. The area under the FROC (AFROC) was used as a summary measure to compare experiments \cite{afroc}. Confidence intervals were also calculated by bootstrapping.

\section{Results}
\label{sec:results}

\subsection{Hard sample mining}

\begin{figure}[!htb]
    \centering
    \includegraphics[width = 0.45\textwidth]{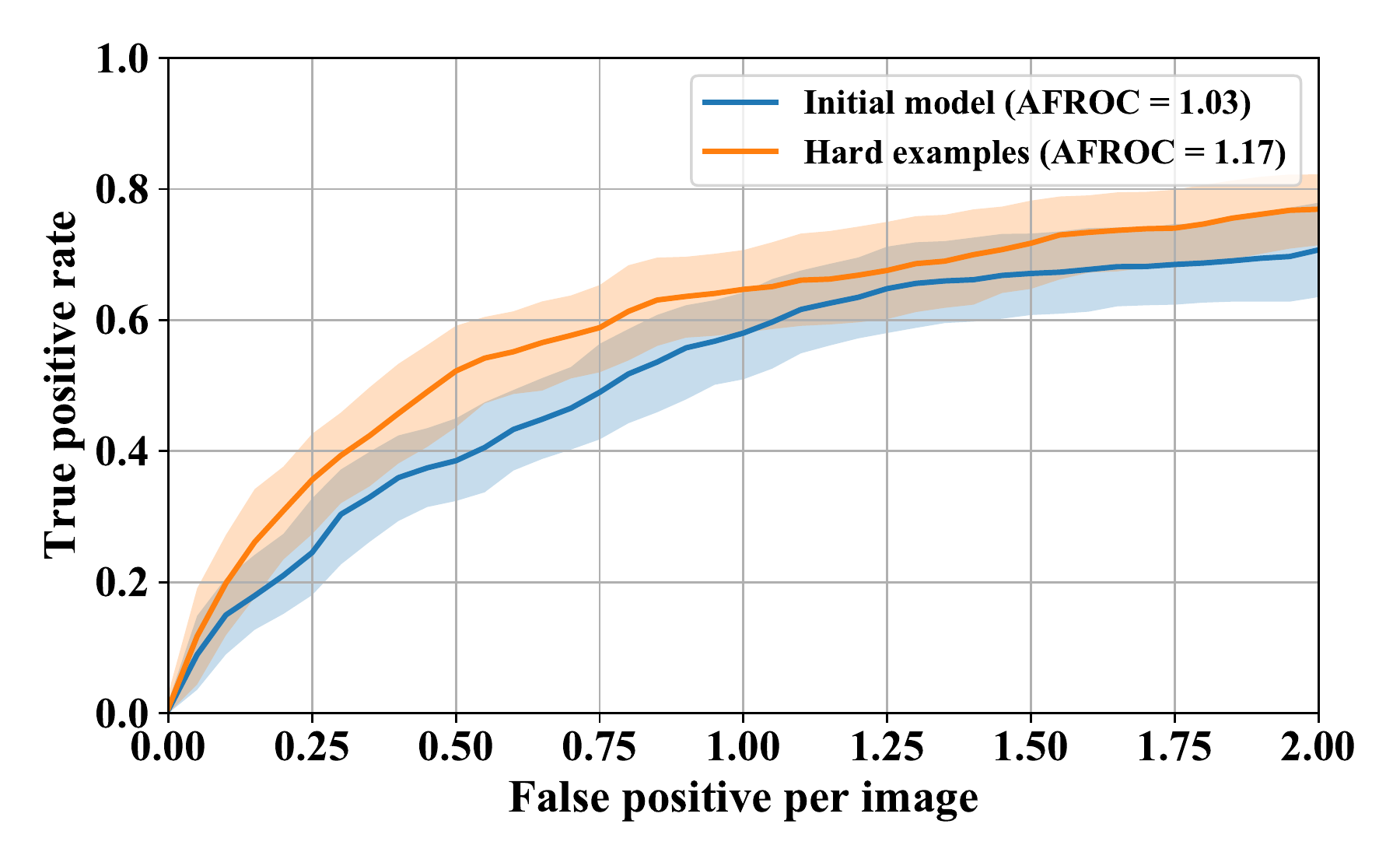}
    \caption{Comparison of the FROC curves with and without applying hard sample mining. }
    \label{froc_clean}
\end{figure}

Introducing hard sample mining improves the AFROC from 1.03 to 1.17 (+0.14), as depicted in Figure \ref{froc_clean}. At 0.5 FPs/image, this corresponds roughly to a 0.15 average increase in sensitivity. Therefore, this procedure was used in all further experiments.

\begin{figure}[tbh]
    \centering
    \includegraphics[width = 0.2\textwidth]{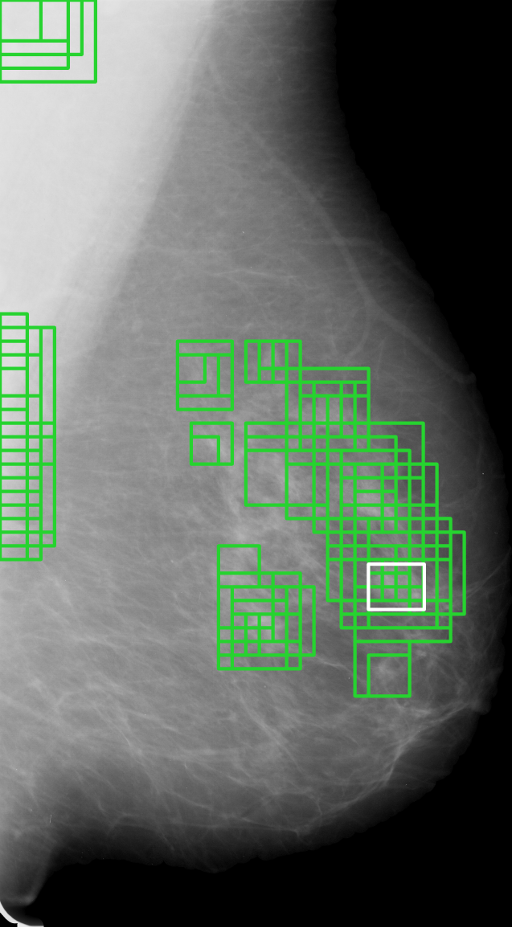}
    \includegraphics[width = 0.2\textwidth]{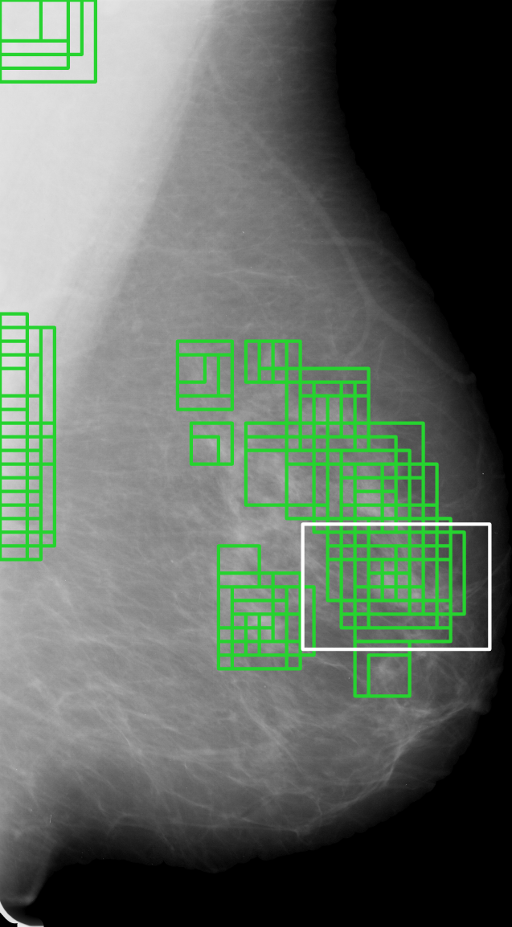}
    \caption{Clean (left) and noisy (right) ground truth box compared with ROI proposals generated by the RPN (shown in green). Many FPs overlap with the noisy ground truth box and, hence, may be incorrectly labeled as foreground.}
    \label{match_example_2}
\end{figure}

\begin{figure*}[!htb]
    \centering
    \includegraphics[width = 1\textwidth]{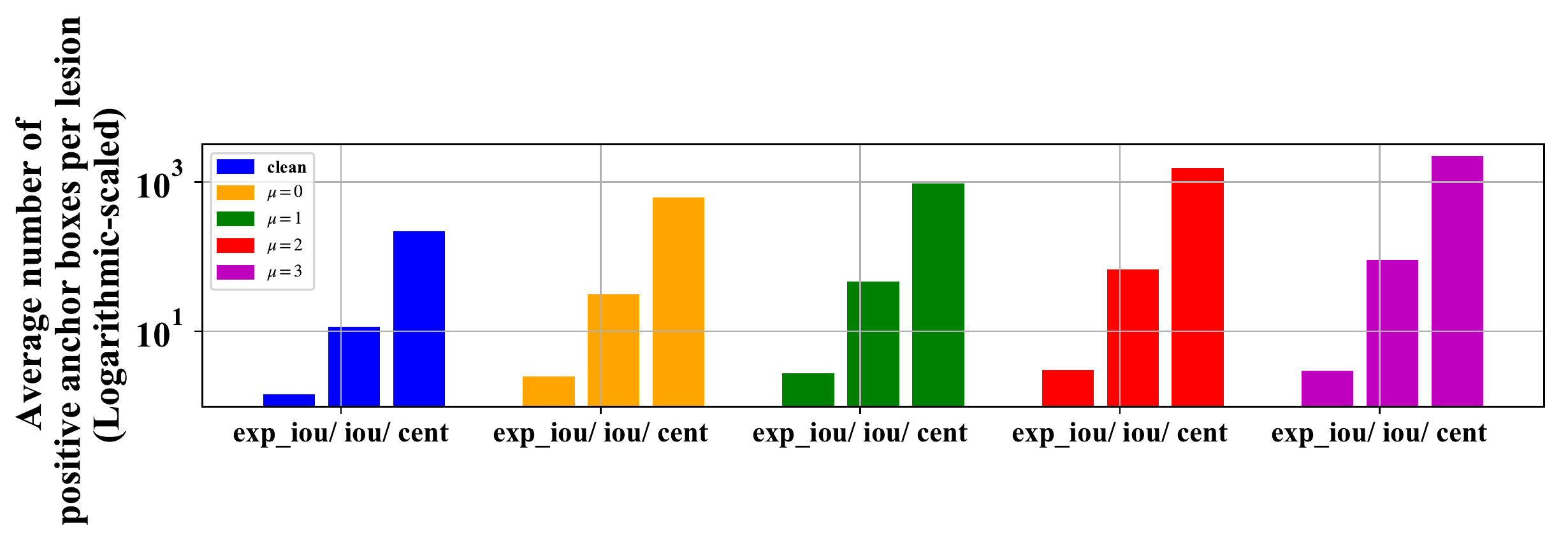}
    \caption{Average number of anchors per lesion labeled as positive in the first iteration of the RPN training.  Results are compared for the Exp\_IoU, IoU and Centroid-based criteria for the clean dataset (blue) and for increasing levels of noise (from yellow to purple). The number of positive anchors (and hence the noise) increases with more relaxed matching criteria and increases more than linearly with the  amount of noise. All scales are logarithmic.}
    \label{noiseperles}
\end{figure*}

\subsection{Effect of matching criteria and noise on bounding box labeling}
In order to achieve a more in-depth understanding of the matching criteria and their tolerance towards noise, we analyzed the average number of anchors  per lesion that were labelled as positive during the first iteration of the RPN training. Given that the actual number of lesions is constant, we assume that an increase in the number of positive anchors  corresponds to labelling noise due to the imperfect ground truth and/or imperfect matching criterion; in fact, as shown in Figure \ref{match_example_2}, many FP ROIs overlap with noisy ground truth boxes and may be incorrectly labelled.  
The results are shown in Figure \ref{noiseperles} for both the clean and noisy datasets. A difference between the number of positive anchors is noticeable across different matching criteria, especially for the centroid criterion which is the most loose. It is also worth noticing that the amount of box coordinate noise results in a higher amount of labelling noise while training the RPN and the detector. 
The number of positive anchors increased up to eight times for the IoU and up to 10 times for the centroid-based criterion. On the contrary, for the proposed Exp\_IoU criterion, the increase is only two-fold, and hence we hypothesize that the final performance will be less sensitive to noise.

\begin{figure*}[htb]
    \centering
    \includegraphics[width = 1\textwidth]{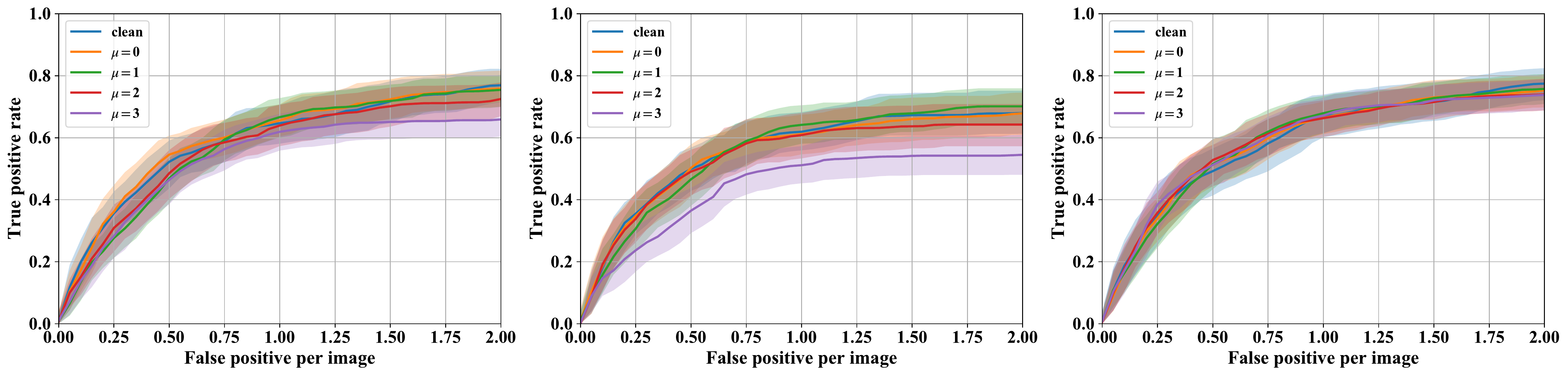}
    \caption{FROC curves with 95\% confidence interval (calculated by bootstrapping). From left to right the IoU, Centroid, and Exp\_IoU criterion were used. The latter is more tolerant towards noise with comparable performance across all levels of noise.}
    \label{ci_boot}
\end{figure*}

\subsection{FROC performance}

\begin{figure}[!htb]
    \centering
    \includegraphics[width = 0.45\textwidth]{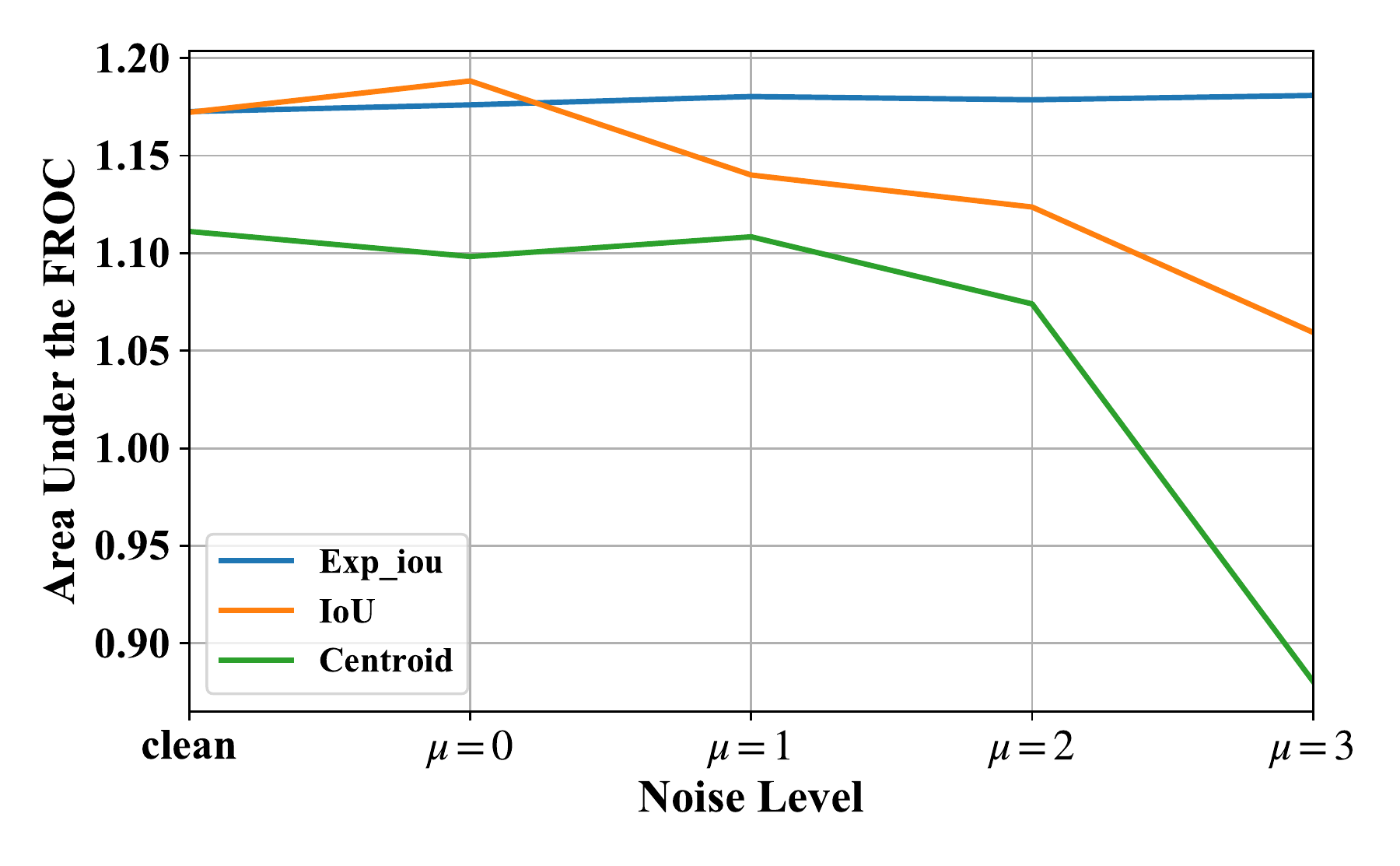}
    \caption{Area under the FROC curves for the IoU (orange), the Centroid inside the bounding box (green) and the Exp\_IoU criteria as a function of the noise level.}
    \label{afroc}
\end{figure}

The FROC curves for different noise levels and matching criteria are shown in Figure \ref{ci_boot}. For an easier comparison, we report the mean AFROC with respect to the noise level in Figure \ref{afroc} and the corresponding confidence intervals in Table \ref{afroc_int}. The results confirm that the Centroid-based criterion is a poor choice for training, since the AFROC is always lower. The best results are achieved by both the IoU and the Exp\_IoU on the clean dataset; however, for the IoU criterion the performance degrades almost linearly from 1.17 to 1.06 with increasing levels of noise. Exp\_IoU is the most robust criterion with respect to noise, as no performance drop can be noticed.

By comparing Figure \ref{noiseperles} and Figure \ref{afroc}, it can be noticed that the number of positive anchors and the performance are negatively correlated, which we attribute to a noisy ground truth degrading the performance of the network. Similar trends have been observed in classification networks where the amount of clean labels is relative small and the ratio of noisy to clean labels exceeds 10:1 \cite{RobustmasslabelnoiseRolnick}.

\begin{table}[!htb]
\renewcommand{\arraystretch}{1.3}
\caption{Confidence intervals for the AFROC.}
\label{afroc_int}
\centering

\scriptsize
\begin{tabular}{|c||c||c||c||c||c|}
\hline
 %\textbf{Criterion} & \textbf{Clean} & \textbf{$\mu =0 $} & \textbf{$ \mu =1 $} & \textbf{$\mu =2 $} & \textbf{$\mu =3 $} \\
 {Criterion} & {Clean} & {$\mu =0 $} & {$ \mu =1 $} & {$\mu =2 $} & {$\mu =3 $} \\
\hline
IoU &1.29-1.05&1.29-1.09&1.23-1.039&1.23-1.01 & 1.16-0.95\\
\hline
Centroid & 1.23-0.9&1.22-0.98&1.23-1.00&1.20-0.94 & 1.00-0.76\\
\hline
Exp\_IoU &1.30-1.04&1.27-1.07& 1.29-1.07&1.28-1.08& 1.28-1.08\\
\hline
\end{tabular}
\end{table}

\begin{figure*}[!htb]
    \centering
    \includegraphics[width=0.6\textwidth]{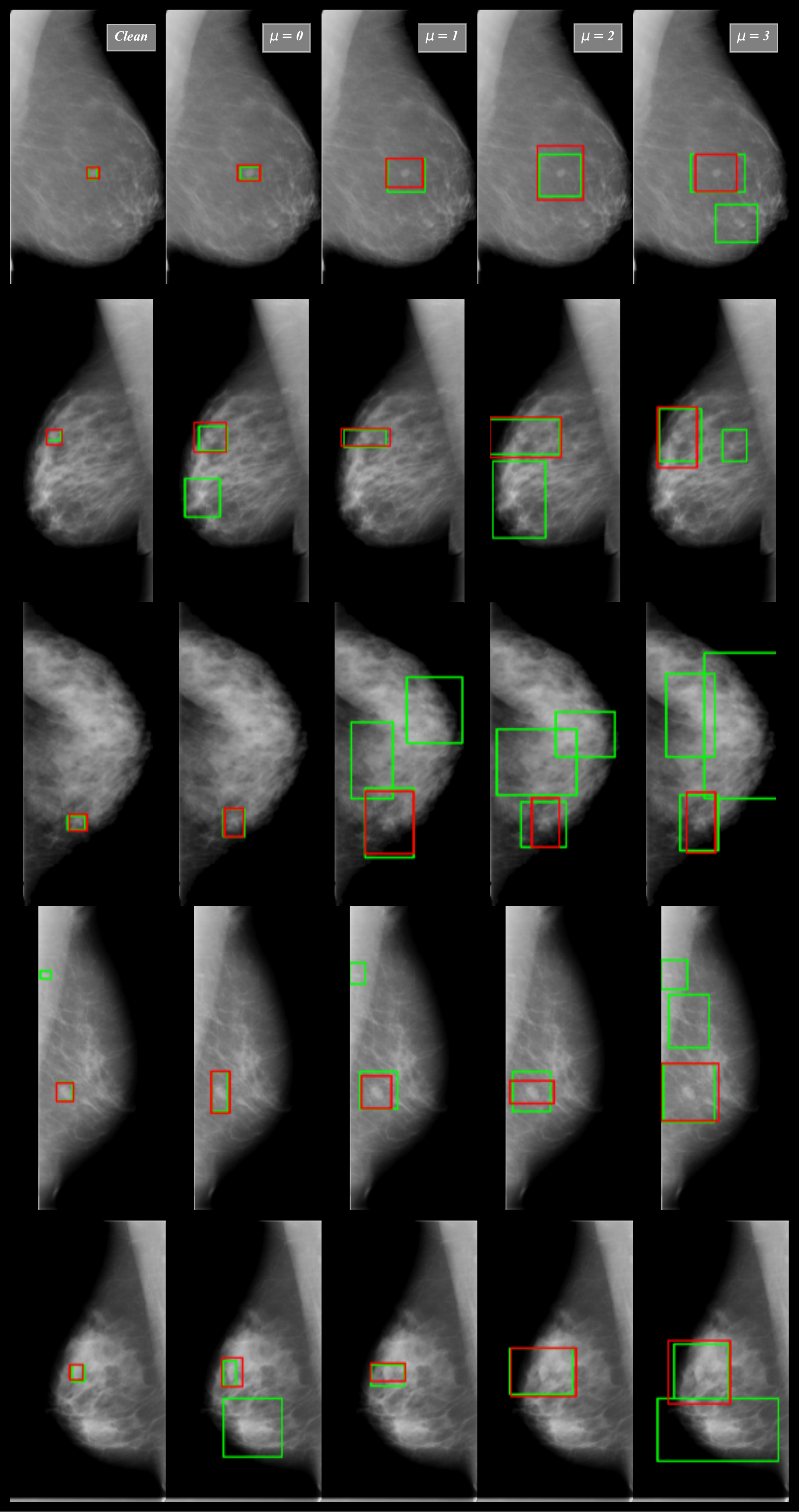}
    \caption{ \hl{Examples showing the clean and noisy ground truth annotations (red) vs. the network predictions (green) in the training set. From left to right the level of noise increases. The matching criterion used was the IoU. It can be seen that, as the noise level increases, additional false positives are produced by the network.}}
    \label{noise_ex}
\end{figure*}

\section{Discussion}
\label{discussion}
In this work, we analyzed a specific type of noise (bounding box coordinate noise), which is extremely relevant to training object detectors from imperfect ground truth and was not covered by the previous literature.  
Our experiments show that coordinate noise, specifically in the form of \textit{enlarged ground truth bounding boxes}, will result in labeling noise while training the classifier heads and, hence ultimately decrease the network performance. 

The network tolerance to noise is mediated by the matching criterion used to label anchor boxes and RPN proposals during training. The standard criterion used in object detection, the IoU, is surprisingly robust to moderate levels of noise ($\mu=1$) and degrades only for very high levels of noise: for instance, at 2 FPs/image the average sensitivity drops from 77\% to 65\%. Other criteria, such as in particular the Exp\_IoU, are able to maintain performance stable across all levels of noise. The main reason is that it explicitly considers the position of the anchor box with respect to the lesion center, and not only the overlapping area. Therefore,  misleading anchor boxes that are not well positioned  will be discarded during training. However, this advantage may be lost if the lesion centers are affected by random noise as well.  

Previous works had already shown that training object detectors for mammography, and medical images in general, requires specific choices of hyper-parameters, such as appropriate IoU thresholds, since lesions are less frequent and with more ill-defined borders compared to objects in natural scenes \cite{ribli2018detecting}. We confirmed such findings and employed a hard negative mining strategy to select informative and balanced samples for training the detector.

Our study has potential limitations. First of all, the dataset size is relatively small, and is not sufficient to reach state-of-the-art performance such as in \cite{ribli2018detecting}. This is partly due to overfitting, which was also observed in previous works \cite{cha2019reducing}. In our experiments, overfitting was mitigated by early stopping, as well as by the hard sample mining procedure. Investigating to which extent overfitting is reduced with larger dataset sizes is an interesting avenue for future work. The impact of noise also depends on the size of the dataset. It is conceivable that, on larger datasets, the noise tolerance could further improve as more examples of lesions become available \cite{RobustmasslabelnoiseRolnick}.

Secondly, with our current training procedure the effect of noise on the regression parameter is unavoidable, i.e., the predicted bounding boxes are going to be enlarged, as shown in Figure \ref{noise_ex}. We leave improving the quality of the regression to future work. For instance, knowing the level of noise in a given dataset, the regression parameters could be adjusted during or after training; strategies to mitigate the effect of bounding box noise on regression were also proposed by Gao and colleagues \cite{gao2018note}.

In this work, noise was simulated by explicitly manipulating annotations in a well-curated dataset. This is a common approach to study the effect of noise in machine learning and allows us to carefully control the experimental conditions. Nonetheless, our conclusions should be validated in a real-life dataset, which we plan on tackling in future work. Likewise, the proposed approach could be extended to other types of lesions or datasets.

\section{Conclusion}

We quantitatively investigated the effect of bounding box coordinate noise while training object detection networks for mammography. We showed how state-of-the-art object detectors are robust to varying degrees of labelling noise, and proposed strategies to mitigate its effect at extreme noise levels. Our study has important implications for dataset collection and annotation, since it shows that the bounding boxes do not need to be very precise for training to be effective. In the case of extreme noise levels, small changes in the training procedure, such as introducing a different matching criterion, can improve performance without increasing the complexity of the model, and can be easily incorporated in the training procedure of any object detector. These findings open new opportunities to train lesion detection model by using bookmarks and annotations routinely recorded by radiologists in their clinical practice.

\section*{Acknowledgments}

We gratefully acknowledge the support of NVIDIA Corporation with the donation of the Titan Xp GPU used for this research.

\bibliographystyle{unsrt}
\bibliography{main}

\newpage

\begin{IEEEbiography}[{\includegraphics[width=1in,height=1.25in,clip,keepaspectratio]{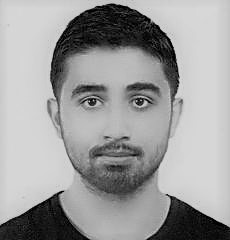}}]{Sina Famouri} has graduated with a master's degree in Artificial Intelligence from Shiraz University, Shiraz, Iran. Currently, he is a Ph.D. student at the Dipartimento di Automatica e Informatica of Politecnico di Torino, Italy, supervised by Prof. Fabrizio Lamberti. His research interests include medical image analysis, computer vision, deep learning, and machine learning. 
\end{IEEEbiography}

\begin{IEEEbiography}[{\includegraphics[width=1in,height=1.25in,clip,keepaspectratio]{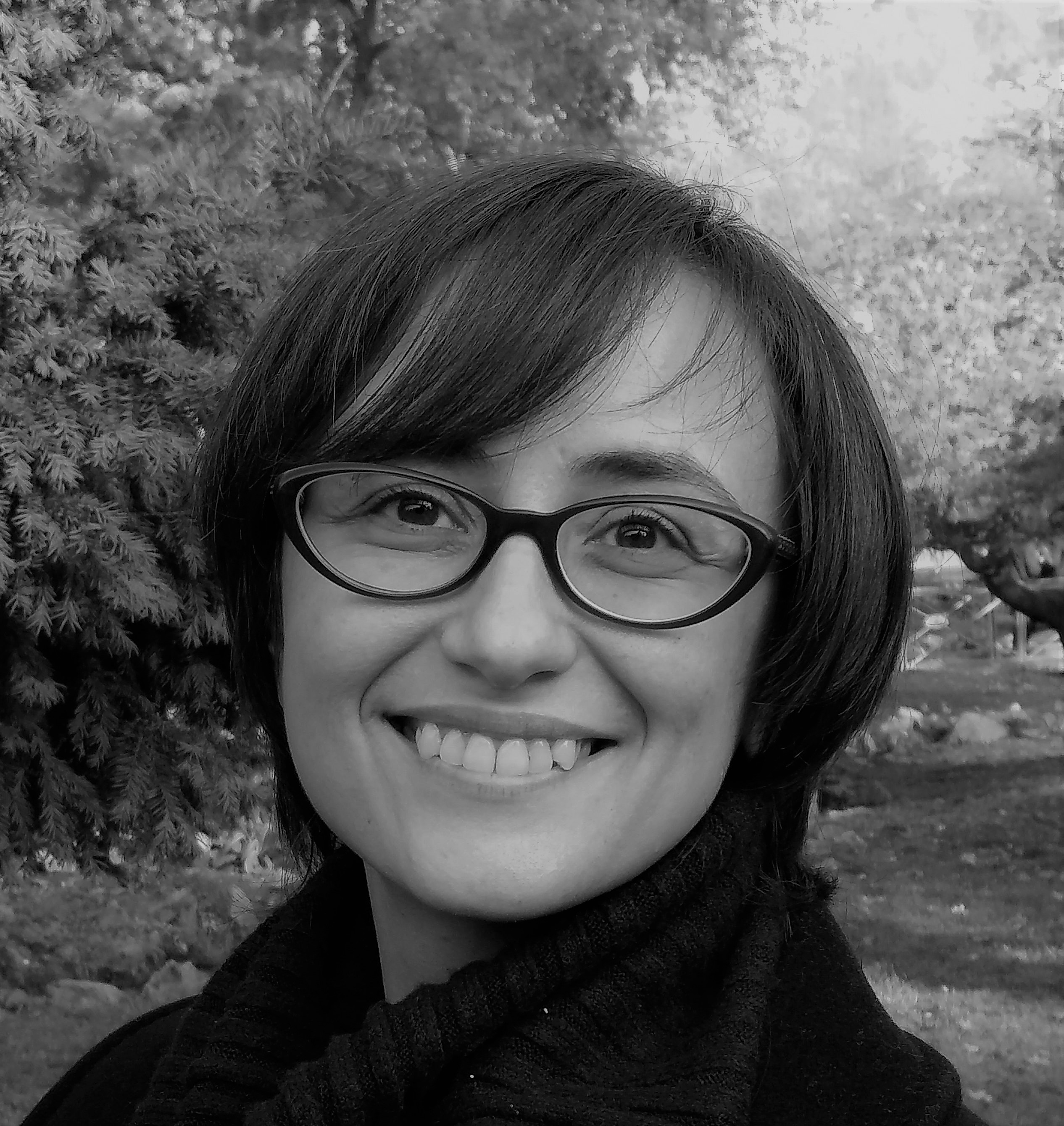}}]{Lia Morra} received the PhD in computer engineering from Politecnico di Torino, Italy, in 2006. Currently, she is senior post-doctoral fellow at the Dip. di Automatica e Informatica of Politecnico di Torino. Her research interests include computer vision, pattern recognition, and machine learning. She is an Associated Editor for IEEE Consumer Electronics Magazine.
\end{IEEEbiography}

\begin{IEEEbiography}[{\includegraphics[width=1in,height=1.25in,clip,keepaspectratio]{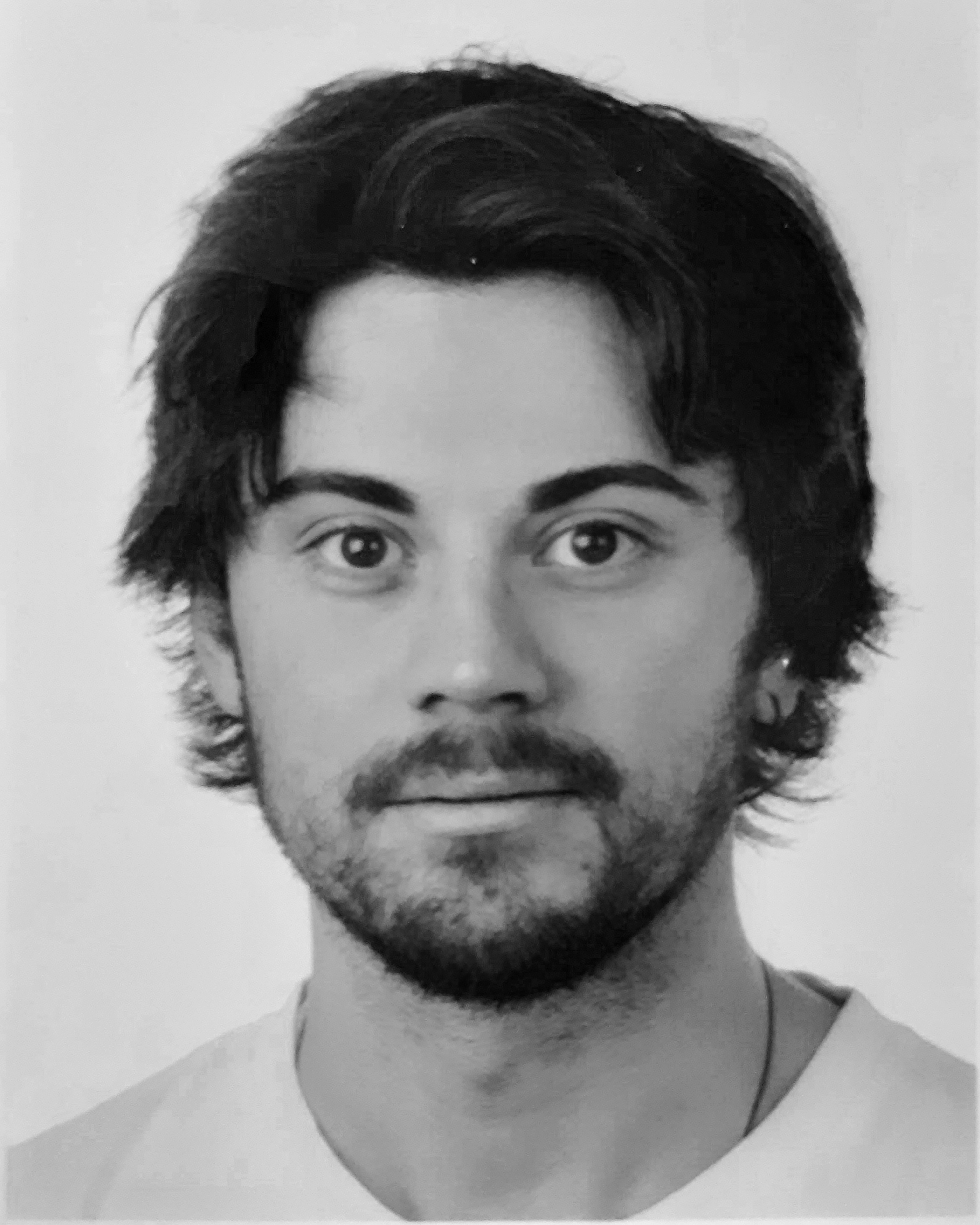}}]{Leonardo Mangia} graduate in biomedical engineering at Politecnico di Torino, Italy, in 2019
with a specialization in designing medical devices. His research interests include machine
learning and computer vision for medical applications. Currently, he’s working as a
biomedical engineer in developing countries to improve their healthcare systems and their
life-conditions.
\end{IEEEbiography}

\begin{IEEEbiography}[{\includegraphics[width=1in,height=1.25in,clip,keepaspectratio]{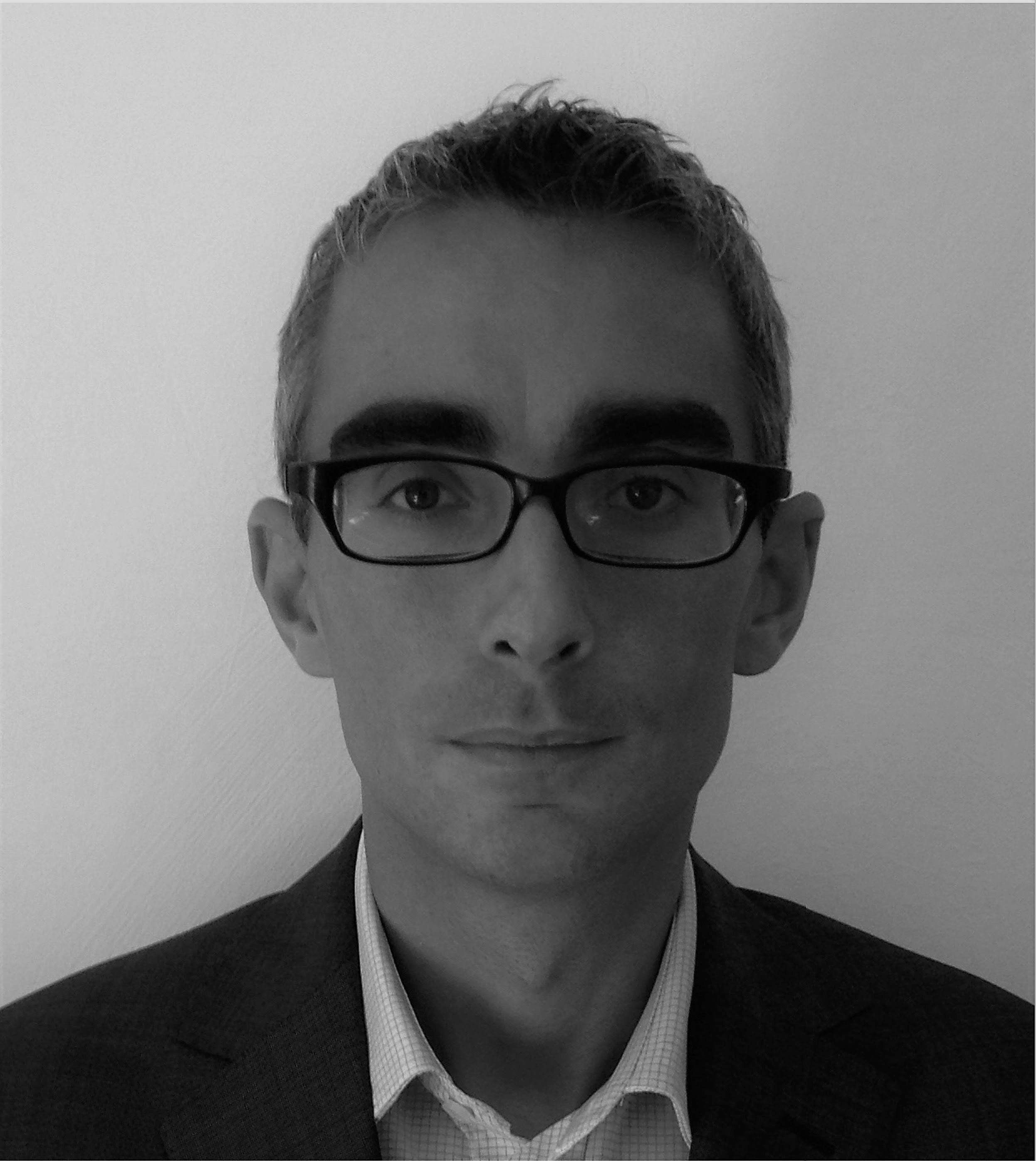}}]{Fabrizio Lamberti} is a Full Professor with the Dipartimento di Automatica e Informatica of Politecnico di Torino, Turin, Italy. His research interests include computer graphics, computer vision, human-machine interaction, and intelligent computing. He is serving as an Associate Editor for IEEE Transactions on Computers, IEEE Transactions on Learning Technologies, IEEE Transactions on Consumer Electronics, IEEE Consumer Electronics Magazine, and the International Journal of Human-Computer Studies.
\end{IEEEbiography}

\EOD

\end{document}